\documentclass{article}

\usepackage{microtype}
\usepackage{graphicx}
\usepackage{subfigure}
\usepackage{booktabs} 

\usepackage{hyperref}

\usepackage{tikz,graphicx,subcaption}
\usetikzlibrary{positioning,fit,arrows.meta,backgrounds,calc}

\usepackage[accepted]{icml2025}

\usepackage{amsmath}
\usepackage{amssymb}
\usepackage{mathtools}
\usepackage{amsthm}

\usepackage[capitalize,noabbrev]{cleveref}

\theoremstyle{plain}

\theoremstyle{definition}

\theoremstyle{remark}

\usepackage[textsize=tiny]{todonotes}

\begin{document}

\twocolumn[

\icmltitle{Physics informed Transformer‑VAE for biophysical parameter estimation: PROSAIL model inversion in Sentinel‑2 imagery}




\begin{icmlauthorlist}
\icmlauthor{Prince Mensah\thanks{Main author.}}{aimsg}
\icmlauthor{Pelumi Victor Aderinto}{aimssa}
\icmlauthor{Ibrahim Salihu Yusuf}{insta}
\icmlauthor{Arnu Pretorius}{insta}
\end{icmlauthorlist}

\icmlaffiliation{aimsg}{African Institute for Mathematical Sciences (AIMS), Accra, Ghana}
\icmlaffiliation{aimssa}{African Institute for Mathematical Sciences (AIMS), Cape Town, South Africa}
\icmlaffiliation{insta}{InstaDeep, Cape Town, South Africa}

\icmlcorrespondingauthor{Prince Mensah}{prince.mensah@aims.edu.gh}

\icmlkeywords{Machine Learning, ICML}

\vskip 0.3in
]




\begin{abstract}
Accurate retrieval of vegetation biophysical variables from satellite imagery is crucial for ecosystem monitoring and agricultural management. In this work, we propose a physics-informed Transformer-VAE architecture to invert the PROSAIL radiative transfer model for simultaneous estimation of key canopy parameters from Sentinel-2 data. Unlike previous hybrid approaches that require real satellite images for self-supevised training. Our model is trained exclusively on simulated data, yet achieves performance on par with state-of-the-art methods that utilize real imagery. The Transformer-VAE incorporates the PROSAIL model as a differentiable physical decoder, ensuring that inferred latent variables correspond to physically plausible leaf and canopy properties. We demonstrate retrieval of leaf area index (LAI) and canopy chlorophyll content (CCC) on real-world field datasets (FRM4Veg and BelSAR) with accuracy comparable to models trained with real Sentinel-2 data. Our method requires no in-situ labels or calibration on real images, offering a cost-effective and self-supervised solution for global vegetation monitoring.	The proposed approach illustrates how integrating physical models with advanced deep networks can improve the inversion of RTMs, opening new prospects for large-scale, physically-constrained remote sensing of vegetation traits.
\end{abstract}

\section{Introduction}
\label{introduction}

Global warming and population growth are driving the need for frequent, high-resolution mapping of vegetation biophysical variables (BV) such as leaf area index (LAI) and canopy chlorophyll content (CCC). LAI (half the total leaf area per ground area) and CCC (total leaf chlorophyll per ground area) are essential for assessing crop health, carbon stocks, and ecosystem productivity. Traditionally, these variables are retrieved from remote sensing reflectance by inverting physical radiative transfer models or using empirical relationships~\cite{jacquemoud1990prospect}. Physical model inversion (e.g., PROSAIL) via look-up tables or optimization can be accurate but is computationally intensive and non-unique (different parameter sets can fit the same spectra)~\cite{zhu2018look,combal2003retrieval,atzberger2004object}. Machine learning (ML) offers a faster alternative by learning mappings from reflectance to biophysical parameters~\cite{houborg2018hybrid,srinet2019estimating}, but standard supervised approaches require large labeled datasets, which are scarce for satellite-derived BV.

A common strategy is to train ML regressors on synthetic data generated by radiative transfer models, then apply them to satellite images. However, simulation-trained ML models often struggle to generalize due to distribution mismatch between simulated and real data~\cite{zhu2018look,combal2003retrieval,atzberger2004object}. The reflectance simulations may not capture all real-world factors (e.g., soil/background variability, canopy structure, sensor noise), and the assumed distributions of input parameters may differ from reality. These issues can lead to biasead predictions when models trained purely on simulation are applied to actual imagery.

To bridge this gap, physics-guided deep learning approaches have emerged. One promising paradigm is to integrate the radiative transfer model itself into a neural network architecture, enabling self-supervised training on unlabeled satellite data~\cite{zerah2024physics}. \cite{zerah2024physics} introduced a PROSAIL-VAE that incorporates a differentiable PROSAIL model as the decoder of a variational autoencoder (VAE). In their hybrid approach, the network is trained directly on real Sentinel-2 reflectance images (without requiring ground-truth LAI/CCC) by forcing the VAE to reconstruct image spectra through the physical model. This self-supervised strategy yielded simultaneous probabilistic inversion of all PROSAIL parameters and achieved accuracy rivaling
conventional methods like the Sentinel-2 Biophysical Processor~\cite{weiss2020s2toolbox}. However, the hybrid training requires a large corpus of satellite imagery and a complex training procedure such as sampling latent variables and Monte Carlo integration.

In this paper, we investigate whether a carefully designed physics-informed model can achieve similar performance using only simulated training data, eliminating the need for any real images in training. We present a Transformer-VAE architecture that embeds the PROSAIL radiative transfer model as a fixed decoder and uses a Transformer-based encoder to infer the distributions of biophysical parameters from input spectra. By training this network end-to-end on a generated dataset of PROSAIL simulations, we leverage physical consistency to generalize well to real-world data. Our approach distinguishes itself by relying solely on synthetic data for training. This demonstrates that, given appropriate simulation realism and a physics-informed network, the simulation-to-reality gap can be largely closed.

The remainder of this article is organized as follows: section \eqref{methodology} describes the methodology, including the PROSAIL model \eqref{prosailmodel}, the Transformer-VAE architecture \eqref{transformer}, and the self-supervised regression framework \eqref{selfsupervised}. Section \eqref{data} details the simulated training dataset and the real field datasets (FRM4Veg and BelSAR) used for evaluation. Section \eqref{results} presents the results for LAI and CCC retrieval, with comparisons to state-of-the-art methods. We discuss the implications and novelty of our simulation-only training approach in Section \eqref{discussion}, and conclude in Section \eqref{conclusion}.

\section{Methodology}
\label{methodology}
\subsection{PROSAIL Radiative Transfer Model}
\label{prosailmodel}
PROSAIL is a widely used radiative transfer model that combines the PROSPECT leaf optical properties model with the SAIL canopy bidirectional reflectance model. PROSPECT simulates the spectral reflectance and transmittance of a representative leaf across the solar-visible and near-infrared spectrum (400–2500 nm) as a function of leaf biochemical and structural parameters~\cite{jacquemoud2009prospect+}.

The Scattering by Arbitrarily Inclined Leaves (SAIL) model uses canopy-level parameters to simulate the reflectance of a vegetation canopy given the single-leaf spectra from PROSPECT. By combining PROSPECT and SAIL, PROSAIL can generate top-of-canopy reflectance spectra for a vegetated scene under specified viewing and illumRination angles~r\cite{verhoef1984light,jacquemoud2009prospect+}.

In this study, we used the PROSAIL to generate continuous reflectance spectra from 400–2500 nm. To simulate Sentinel-2 observations, we convolve the output spectra with Sentinel-2 band spectral response functions, producing reflectances for the 10 Sentinel-2 bands in the 443–2190 nm range (excluding the 940 nm water vapor band and SWIR cirrus band)~\cite{zerah2024physics}.  We include the sensor viewing geometry in the simulations: each reflectance sample is generated for a random solar zenith angle (e.g. 15\textdegree–60\textdegree), sensor zenith (0\textdegree–10\textdegree off-nadir), and sun-sensor azimuth difference (0\textdegree–180\textdegree). These angles are provided as additional inputs to the model during inversion, as detailed below in \eqref{transformer} . By using a consistent physical model for both data generation and inversion, we ensure that our neural network’s task is physically well-defined: any reflectance it encounters (simulated or real) is assumed to be explainable by some combination of PROSAIL parameters within realistic bounds.

\subsection{Transformer-VAE Architecture}
\label{transformer}
\begin{figure*}[t]
	\centering
	
	\resizebox{\textwidth}{!}{%
		\begin{tikzpicture}[
			data/.style       = {draw, rectangle, inner sep=1pt, align=center},
			block/.style      = {draw, rectangle, rounded corners, align=center,
				minimum width=2.5cm, minimum height=1cm},
			encoder/.style    = {block, fill=purple!20},
			posenc/.style     = {block, fill=orange!20},
			latent/.style     = {block, fill=yellow!20},
			prosaildec/.style = {block, fill=orange!20},
			prosailvar/.style = {block, fill=white},
			loss/.style       = {draw=red!80!black, thick, rounded corners, align=center,
				minimum width=1.8cm, minimum height=0.8cm},
			arrow/.style      = {thick, ->, >=Stealth},
			dashedarrow/.style= {arrow, dashed},
			dottedarrow/.style= {arrow, dotted}
			]
			
			\node[block] (s2img) {\includegraphics[width=1.8cm]{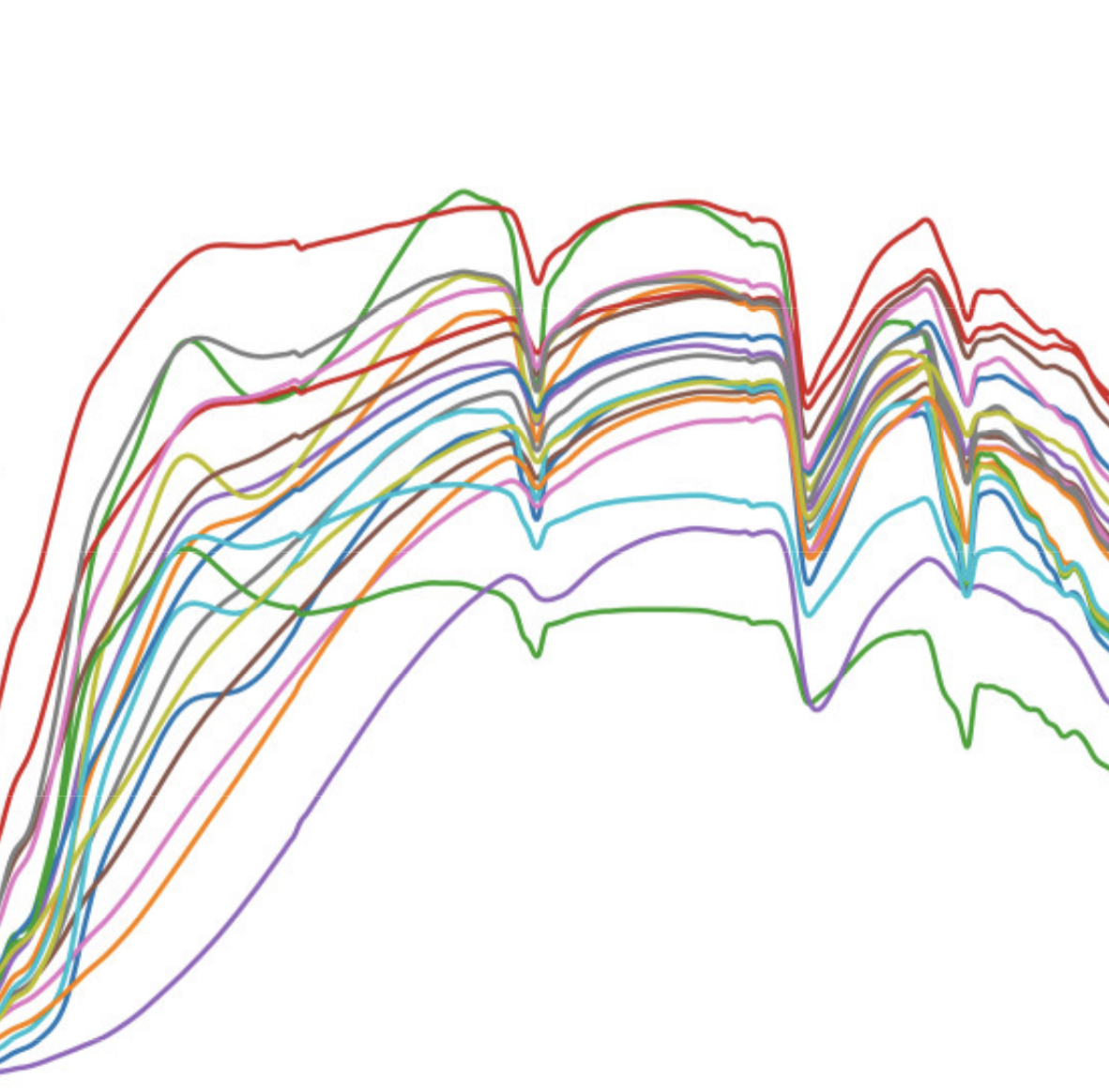}\\S2 simulated\\reflectance};
			\node[block, below=25mm of s2img] (angles) {Viewing\\angles};
			
			\node[posenc, right=20mm of s2img] (posenc) {Positional\\Encoding};
			\draw[arrow] (s2img) -- (posenc);
			\draw[arrow] (angles.east) to[out=0,in=180] (posenc.west);
			
			\node[encoder, right=of posenc] (transenc) {Transformer\\Encoder};
			\draw[arrow] (posenc) -- (transenc);
			
			\node[latent, right=of transenc] (latdist)
			{Latent heads\\$\{\mu_i,\sigma_i\}_{i=1}^N$};
			\draw[arrow] (transenc) -- (latdist);
			
			\node[block, right=of latdist] (latsamp)
			{Sample $z_i\sim\mathrm{TruncNormal}(\mu_i,\sigma_i)$};
			\draw[dashedarrow] (latdist.east) --
			node[above=30pt,midway,fill=white]{Reparameterization trick}
			(latsamp.west);
			
			\node[loss, below=of latdist] (kl) {$\mathcal L_{KL}$};
			\draw[dottedarrow] (latdist.south) -- (kl.north);
			
			\node[prosailvar, right=25mm of latsamp] (provars) {PROSAIL\\inputs};
			\draw[arrow] (latsamp) -- node[above]{scale\,\&\,clip} (provars);
			
			\node[prosaildec, right=15mm of provars, yshift=20mm] (prospect) {PROSPECT 5};
			\node[block, below=of prospect] (leaf) {Leaf\\reflectance};
			\node[prosaildec, below=of leaf] (sail4) {4SAIL};
			\draw[arrow] (prospect) -- (leaf);
			\draw[arrow] (leaf) -- (sail4);
			
			\draw[arrow] (angles.east)
			-- ++(0,4mm) -- ++(0,-5mm) -| (sail4.south);
			
			\node[block, right=20mm of sail4] (canopy) {Canopy\\reflectance};
			\node[prosaildec, above=of canopy] (sensor) {S2 sensor\\model};
			\draw[arrow] (sail4) -- (canopy);
			\draw[arrow] (canopy) -- (sensor);
			
			\begin{scope}[on background layer]
				\node[draw, thick, inner sep=6pt,
				fit=(prospect)(leaf)(sail4)(canopy)(sensor)] (decoder) {};
				\node[above=1mm of decoder.north]
				{Decoder:\,\textbf{PROSAIL}};
			\end{scope}
			
			\draw[arrow] (provars.north)
			to[out=90,in=180] (prospect.west);
			\draw[arrow] (provars.south)
			to[out=-90,in=180] (sail4.west);
			
			\node[data, right=of sensor] (recon)
			{\includegraphics[width=1.8cm]{figs/sim_refl}\\Reconstructed\\reflectance};
			\draw[arrow] (sensor) -- (recon);
			
			\node[loss] (lrec)
			at($ (s2img)!0.5!(recon)+(0,40mm)$) {$\mathcal L_{rec}$};
			\draw[dottedarrow] (recon.north) |- (lrec.east);
			\draw[dottedarrow] (s2img.north) |- (lrec.west);
			
		\end{tikzpicture}
	}
	\\
	\vspace{0.5em}
	{\small (a) Training Phase}
	
	\bigskip
	
	\resizebox{\textwidth}{!}{%
		\begin{tikzpicture}[
			data/.style       = {draw, rectangle, inner sep=1pt, align=center},
			block/.style      = {draw, rectangle, rounded corners, align=center,
				minimum width=2.5cm, minimum height=1cm},
			encoder/.style    = {block, fill=purple!20},
			posenc/.style     = {block, fill=orange!20},
			latent/.style     = {block, fill=yellow!20},
			prosailvar/.style = {block, fill=white},
			arrow/.style      = {thick, ->, >=Stealth}
			]
			
			\node[block] (s2img) {\includegraphics[width=1.8cm]{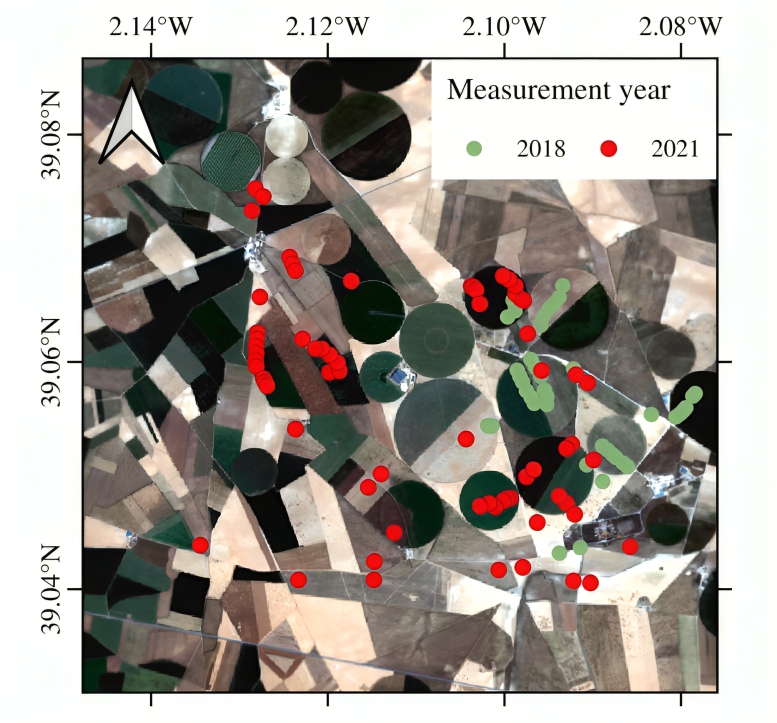}\\S2 image};
			\node[block, below=25mm of s2img] (angles) {Sun/S2\\angles};
			
			\node[posenc, right=20mm of s2img] (posenc) {Positional\\Encoding};
			\draw[arrow] (s2img) -- (posenc);
			\draw[arrow] (angles.east) to[out=0,in=180] (posenc.west);
			
			\node[encoder, right=of posenc] (transenc) {Transformer\\Encoder};
			\draw[arrow] (posenc) -- (transenc);
			
			\node[latent, right=of transenc] (latdist)
			{Latent heads\\$\{\mu_i,\sigma_i\}_{i=1}^N$};
			\draw[arrow] (transenc) -- (latdist);
			
			\node[prosailvar, right=25mm of latdist] (provars) {PROSAIL\\parameter\\distributions};
			\draw[arrow] (latdist.east)
			-- node[above]{scale\,\&\,clip} (provars);
			
			\node[data, right=25mm of provars, yshift=25mm] (estimates)
			{\includegraphics[width=1.8cm]{figs/barrax}\\Parameter estimates};
			\draw[arrow] (provars.north)
			to[out=90,in=180] (estimates.west);
			
			\node[data, right=25mm of provars, yshift=-25mm] (uncertainties)
			{\includegraphics[width=1.8cm]{figs/barrax}\\Parameter uncertainties};
			\draw[arrow] (provars.south)
			to[out=-90,in=180] (uncertainties.west);
			
			\node[block, right=40mm of $(estimates)!0.5!(uncertainties)$] (validation)
			{Validation};
			\node[block, fill=green!20, right=25mm of validation] (insitu)
			{In-situ\\data};
			
			\draw[arrow] (insitu.west) -- (validation.east);
			\draw[arrow] (estimates.east)
			to[out=0,in=90]  (validation.north);
			\draw[arrow] (uncertainties.east)
			to[out=0,in=-90] (validation.south);
			
		\end{tikzpicture}
	}
	\\
	\vspace{0.5em}
	{\small (b) Inference Phase}
	
	\caption{End‐to‐end architecture: (a) training with transformer-VAE \& PROSAIL decoder, (b) inference and validation.}
	\label{fig:architecture}
\end{figure*}

We frame the inversion of PROSAIL as a variational autoencoder (VAE) problem, where the goal is to encode an input reflectance spectrum into a distribution over physical parameters that can reconstruct the input via the PROSAIL decoder. Our proposed Transformer-VAE architecture in figure~\eqref{fig:architecture} consists of: (1) an encoder network with a Transformer-based architecture that produces a latent probability distribution for each PROSAIL input variable, and (2) a decoder that is the PROSAIL model itself, implemented as a deterministic differentiable function with no trainable weights. The overall architecture ensures that encoding–decoding a spectrum corresponds to finding a physically consistent set of parameters that reproduce the spectrum.

\subsubsection{Encoder}
The encoder takes as input a feature vector representing a pixel’s reflectance and observation angles. Specifically, we use the Sentinel-2 reflectance in 10 bands (blue to SWIR) and three angular features (solar zenith, view zenith, relative azimuth). We adopt a Transformer-based encoder to effectively model relationships among spectral bands and between spectral and angular features. The input is first embedded (each band and angles treated as a token with positional encoding) and passed through multi-head self-attention layers. This allows the encoder to capture complex spectral correlations (e.g., how variations in red and NIR bands relate to LAI or chlorophyll) and to account for angular effects modulating the spectrum. The Transformer encoder yields an output feature vector which is then projected to the parameters of the latent distributions for each target variable. In our VAE, we assume each PROSAIL variable $V$ follows a two-sided truncated normal (TN) distribution within a physically valid range~\cite{zerah2024physics}. The encoder outputs the mean $\mu_V$ and standard deviation $\sigma_V$ of each variable’s distribution (in an unconstrained space), which are then transformed to enforce the truncation bounds (via a scaled logistic function to map to $[\text{min}, \text{max}]$ for that variable). In this way, the encoder explicitly outputs a probabilistic estimate of each biophysical variable, rather than a single point estimate.

\subsubsection{Decoder}
The decoder is the fixed PROSAIL model. During the forward pass, a sample is drawn from each latent TN distribution output by the encoder. These sampled values are fed into PROSAIL to generate a reconstructed reflectance spectrum, which is then compared to the input. Because PROSAIL is implemented in a differentiable manner, gradients can be back-propagated from the reconstruction loss through PROSAIL to the encoder parameter. Notably, only the encoder’s weights are trainable; the decoder (PROSAIL) encapsulates prior physical knowledge and does not learn from data~\cite{zerah2024physics}. At inference time, we discard the decoder and use the encoder to predict the posterior distributions of the variables given a new reflectance. The mean of each inferred distribution can serve as the point estimate for that variable, while the variance provides an uncertainty estimate.

By integrating PROSAIL into the network, we enforce that the latent space of the VAE corresponds to actual physical parameters. Unlike a conventional autoencoder, where the decoder might learn an arbitrary mapping~\cite{kingma2013auto}, our decoder is physically interpretable. The Transformer-based encoder further provides flexibility in capturing nonlinear interactions in the spectra. 

\subsection{Self-Supervised Regression Framework}
\label{selfsupervised}
The proposed Transformer-VAE model is trained using a self-supervised regression framework, leveraging only simulated reflectance data and a reconstruction objective. Given an input reflectance spectrum \( \mathbf{x} \), the encoder produces a distribution \( q_\phi(\mathbf{z}|\mathbf{x}) \) over latent variables \( \mathbf{z} = (\text{LAI}, \text{Cab}, \ldots) \). A latent sample \( \mathbf{z} \sim q_\phi(\mathbf{z}|\mathbf{x}) \) is passed through the decoder—implemented via the PROSAIL model—to reconstruct the reflectance \( \hat{\mathbf{x}} \). Training seeks to minimize the discrepancy between \( \mathbf{x} \) and \( \hat{\mathbf{x}} \), while regularizing the latent posterior toward a predefined prior \( p(\mathbf{z}) \).

The total loss function follows the variational autoencoder (VAE) formulation:
\vspace{-0.1cm}
\begin{equation}
	\mathcal{L} = \mathcal{L}_{\text{rec}} + \beta_{\text{KL}} \, \mathcal{L}_{\text{KL}},
	\label{eq:prosailvae_loss}
\end{equation}

where \( \mathcal{L}_{\text{rec}} \) denotes the reconstruction loss and \( \mathcal{L}_{\text{KL}} \) is the Kullback-Leibler (KL) divergence between the approximate posterior and the prior. The hyperparameter \( \beta_{\text{KL}} \) controls the trade-off between reconstruction fidelity and latent space regularization.

The reconstruction term encourages the decoder to accurately replicate the input reflectance from the latent parameters. Assuming a Gaussian likelihood, it is given by:

\begin{align}
	\mathcal{L}_{\text{rec}} =\ 
	& -\mathbb{E}_{q_\phi(\mathbf{z}|\mathbf{x})}[\log p_\theta(\mathbf{x}|\mathbf{z})] \nonumber \\
	& = \frac{1}{2} \sum_{l} \left[ \log 2\pi \sigma^2_{\theta,l}(\mathbf{z}) 
	+ \frac{(x_l - \mu_{\theta,l}(\mathbf{z}))^2}{\sigma^2_{\theta,l}(\mathbf{z})} \right],
	\label{eq:reconstruction_loss}
\end{align}

where \( x_l \) denotes the \( l \)-th spectral band of the input, and \( \mu_{\theta,l}(\mathbf{z}) \), \( \sigma^2_{\theta,l}(\mathbf{z}) \) are the predicted mean and variance of the decoder output.

The KL divergence term regularizes the latent space by enforcing proximity between the learned posterior and the prior:

\begin{equation}
	\mathcal{L}_{\text{KL}} = \text{KL}\left[q_\phi(\mathbf{z}|\mathbf{x}) \,\|\, p(\mathbf{z})\right],
	\label{eq:kld}
\end{equation}

with \( p(\mathbf{z}) \) defined as a product of independent uniform distributions within valid physiological ranges for each parameter~\cite{zerah2023physics}. This choice imposes physically plausible constraints on the inversion process. To improve convergence and generalization, training begins with a small value of \( \beta_{\text{KL}} \), focusing initially on reconstruction quality, and progressively increases it to enforce stronger regularization.

After training, the encoder alone performs inference: for a given reflectance input, it outputs the posterior distribution over the biophysical parameters. The final estimate for each parameter is the mean of the corresponding truncated normal distribution. Since the model is never trained on ground truth parameters for real reflectance data, this constitutes a self-supervised regression framework. The decoder’s physical consistency (via PROSAIL) ensures the encoder learns meaningful inverse mappings from reflectance to biophysical traits.

Model performance is evaluated using several metrics. The root mean square error (RMSE) quantifies the average prediction error:

\begin{equation}
	\text{RMSE} = \sqrt{\frac{1}{n} \sum_{i=1}^n (x_i - \hat{x}_i)^2},
	\label{eq:rmse}
\end{equation}

where \( x_i \) is the ground truth reflectance and \( \hat{x}_i \) is the model prediction. Uncertainty is assessed using the mean prediction interval width (MPIW), defined as:

\begin{equation}
	\text{MPIW} = \frac{1}{n} \sum_{i=1}^n (u_{\hat{x}_i} - l_{\hat{x}_i}),
	\label{eq:mpiw}
\end{equation}

where \( [l_{\hat{x}_i}, u_{\hat{x}_i}] \) denotes the prediction interval for sample \( i \). The prediction interval coverage probability (PICP) measures the proportion of ground truth values that fall within the predicted intervals:

\begin{equation}
	\text{PICP} = \frac{1}{n} \sum_{i=1}^n \mathbb{I}\left\{ x_i \in [l_{\hat{x}_i}, u_{\hat{x}_i}] \right\},
	\label{eq:picp}
\end{equation}

with \( \mathbb{I}\{\cdot\} \) being the indicator function.

The final encoder model contains 798,358 trainable parameters and is highly efficient, enabling rapid pixel-wise inference on a CPU. Importantly, the self-supervised training paradigm eliminates the need for real-world calibration or in-situ parameter labels, enabling fully simulation-driven learning with physically grounded predictions.

\section{Dataset}
\label{data}
\subsection{Range of PROSAIL parameters and distribution.}
The simulated training dataset was designed to cover the plausible range of vegetation and soil conditions while incorporating prior knowledge of variable correlations. Table \eqref{tab:prosail_parameters} shows the input variables that we sampled for the PROSAIL model. Ten variables were sampled from a two-sided truncated normal (TN) distribution within a defined physical range. The truncation bounds were set to match those used by the Sentinel-2 Biophysical Processor’s training LUT \cite{weiss2020s2toolbox} as closely as possible. For example, LAI was allowed to vary from 0 to 10, $Cab$ from 20 to 90 $\mu g$ $cm^{–2}$, etc., consistent with prior studies~\cite{zerah2024physics}. The soil reflectance was parameterized by two variables: $\rho_S$ (spectral shape, representing soil wetness factor/color, ranged 0–1 uniform) and $r_S$ (soil brightness factor, TN distribution 0.3–3.5). Illumination and observation angles were sampled as described in Section~\eqref{prosailmodel}

\begin{table*}[htp]
	\label{tab:prosailvariables}
	\centering
	\renewcommand{\arraystretch}{1.3}
	\setlength{\tabcolsep}{11.5pt}
	\caption{PROSAIL input parameters with their ranges and distributions.}
	\label{tab:prosail_parameters}
	\begin{tabular}{lllcccc}
		\hline
		Model & Input & Description & Distribution & Unit & Min & Max \\
		\hline
		PROSPECT-5 & $N$ & Leaf structure parameter & TN & $-$ & 1.2 & 1.8 \\
		& $C_{ab}$ & Chlorophyll $a + b$ content & TN & $\mu \text{g}~\text{cm}^{-2}$ & 20.0 & 90.0 \\
		& $C_{w}$ & Water equivalent thickness & TN & cm & 0.0075 & 0.0750 \\
		& $C_{c}$ & Carotenoid concentration & TN & $\mu \text{g}~\text{cm}^{-2}$ & 5 & 23 \\
		& $C_{m}$ & Dry matter content & TN & $\text{g}~\text{cm}^{-2}$ & 0.003 & 0.011 \\
		& $C_{b}$ & Brown pigments content & TN & a.u.p.s.u$^{\text{a}}$ & 0 & 2 \\
		4SAIL & LAI & Leaf Area Index & TN & $-$ & 0 & 10 \\
		& $\alpha$ & Mean leaf angle & TN & Deg & 30 & 80 \\
		& $h$ & Hotspot parameter & TN & $-$ & 0.0 & 0.5 \\
		& $\rho_S$ & Soil wetness factor & Uniform & $-$ & 0 & 1 \\
		& $r_S$ & Soil brightness factor & TN & $-$ & 0.3 & 3.5 \\
		Geometry & $\theta_S$ & Solar zenith angle & Uniform & Deg & 15 & 60 \\
		& $\theta_O$ & Observer zenith angle & Uniform & Deg & 0 & 10 \\
		& $\psi_{SO}$ & Relative azimuth angle & Uniform & Deg & 0 & 180 \\
		\hline
		\multicolumn{7}{l}{$^{\text{a}}$ arbitrary unit per surface unit. TN: Truncated Normal distribution.} \\
	\end{tabular}
\end{table*}

To ensure the simulated distributions resemble real-world conditions, we incorporated dependencies between canopy variables. In particular, LAI was treated as a driving variable: for samples with high LAI, we adjusted the means of other distributions to reflect mature, dense canopies (e.g., higher $Cab$, slightly higher $N$, lower soil brightness). Specifically, when LAI was near its upper bound (e.g. LAI ~10 in our sampling, corresponding to very dense crops or forests), the $Cab$ distribution was truncated to 45–90 (thus favoring greener leaves), $N$ was restricted to 1.3–1.8 (slightly higher structure index), and soil parameters were limited to darker values ($r_S$ around 0.5–1.2). These co-distributions, inspired by the approach of \cite{zerah2024physics}, aimed to avoid physically unrealistic combinations (such as extremely high LAI with very low chlorophyll or very bright soil background) that could confuse the network during training. For low LAI scenarios, we allowed broader variation in leaf traits. This strategy effectively embeds a prior that, for example, lush canopies tend to have green healthy leaves and obscure the soil, whereas sparse canopies might have a wide range of leaf chlorophyll and more visible soil. We emphasize that these correlations were modest and used only in data generation; the network is still free to infer any combination within ranges, but it was mainly trained on realistic ones.

The PROSAIL simulations were done at high spectral resolution (1 nm) and then averaged with Sentinel-2 spectral response functions to obtain band reflectances~\cite{tupin2014remote}. We added a small Gaussian noise (~0.5\% reflectance) to each band to mimic sensor noise and atmospheric residual errors. The final training dataset consisted of 500,000 simulated samples, each with 10-band reflectance and 3 angles. This large sample size ensures the VAE experiences a wide variety of spectra, mitigating overfitting to particular patterns. A separate validation set of 20,000 simulations (with parameter values sampled independently but from the same distributions) was generated to monitor training convergence (reconstruction error and KL divergence) and tune the $\beta_{KL}$ schedule.

\subsection{In-situ Field Data}
After training on simulations, we evaluated the model on two field datasets that provide in-situ measurements of LAI and CCC along with concurrent Sentinel-2 observations. No further fine-tuning was done on these datasets; we applied the trained encoder directly to Sentinel-2 reflectance and compared its predictions to ground truth.

\subsubsection{Fiducial Reference Measurements for Vegetation (FRM4Veg)}
\begin{figure}[htp]
	\begin{center}
		\centerline{\includegraphics[width=\columnwidth]{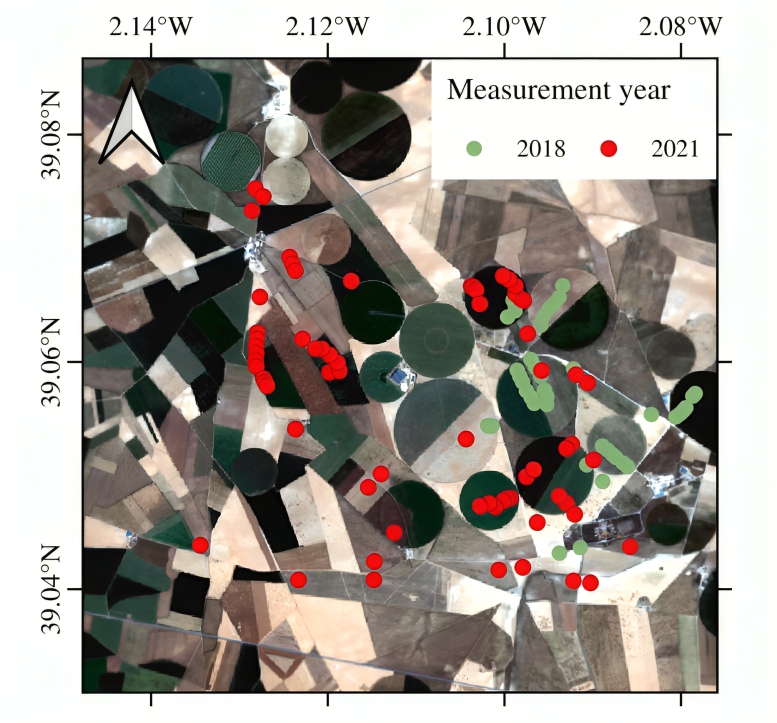}}
		\caption{Spatial distribution of field validation points at the Las Tiesas-Barrax agricultural test site in central Spain, with measurements collected during two separate campaigns in 2018 (green points) and 2021 (red points).}
		\label{fig:barrax}
	\end{center}
	\vskip -0.1in
\end{figure}

The FRM4Veg dataset is part of an ESA project focused on establishing traceable in-situ measurements for validating satellite products[]. It includes data from multiple field campaigns under the FRM4Veg program in 2018 and 2021~\cite{origo2020fiducial, brown2021fiducial}. We specifically use two test sites from FRM4Veg:

\textbf{Las Tiesas Farm (Barrax), Spain:} An agricultural site with irrigated crops. Campaigns were conducted in 2018 and 2021 during the growing season. Ground measurements of LAI were made on various crop fields (wheat, barley, alfalfa, etc.) using standard methods (e.g., LAI-2200 plant canopy analyzer and digital hemispherical photography) on $20 \text{m} \times 20 \text{m}$ Elementary Sampling Units (ESUs). Leaf chlorophyll concentration was measured via destructive sampling or chlorophyll meters (SPAD), enabling computation of CCC (CCC = LAI × leaf chlorophyll content) for many plots. Measurements were accompanied by estimates of uncertainty. We use data from Barrax 2018 and 2021 campaigns, which together provided dozens of ESUs spanning LAI from ~0 (bare soil) to ~6 (dense crops) and CCC from ~0 to ~500 (units: e.g., mg $m^{–2}$). Figure \eqref{fig:barrax} shows a spatial distribution of field validation points at the Las Tiesas-Barrax agricultural test site with measurements collected in 2018 (green points) and 2021 (red points). Following~\cite{brown2021validation}, we exclude a few data points (e.g., an alfalfa field in 2018 that was cut after in-situ measurement but before the satellite overpass leading to mismatch).

\begin{figure}[htp]
	\begin{center}
		\centerline{\includegraphics[width=\columnwidth]{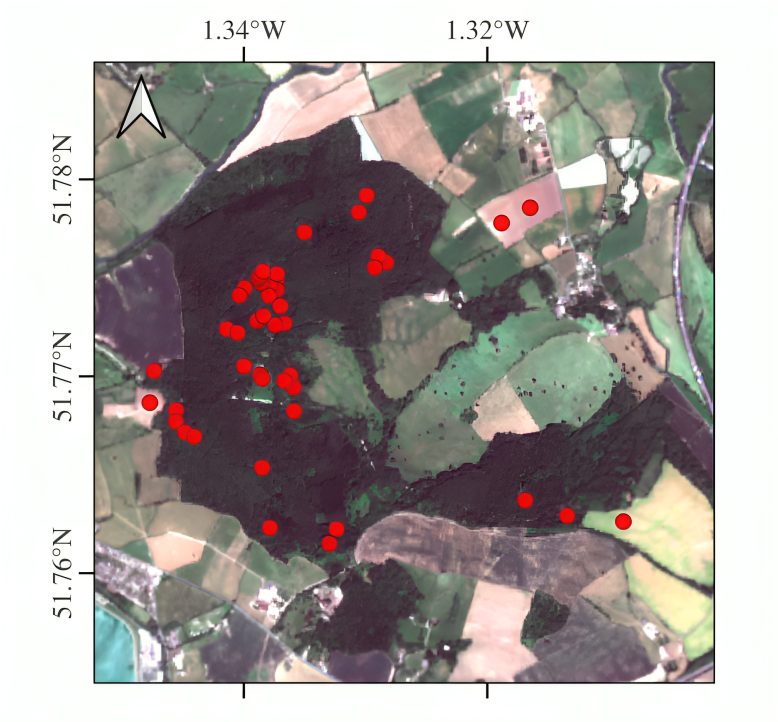}}
		\caption{Field validation points at Wytham Woods, showing the distribution of in-situ LAI and CCC measurements (red points) throughout the ancient deciduous forest site.}
		\label{fig:wytham}
	\end{center}
\end{figure}

\textbf{Wytham Woods, UK:} A deciduous broadleaf forest site. FRM4Veg measurements at Wytham (2018 campaign) include LAI estimated from litterfall and hemispherical photos, and leaf chlorophyll content from sampled leaves (converted to CCC). The forest LAI ranges up to ~5–6, with lower CCC per unit LAI compared to crops (broadleaf chlorophyll ~30–50 $\mu g cm^{–2}$). figure~\eqref{fig:wytham} shows field validation points at Wytham Woods, showing the distribution of in-situ LAI and CCC measurements (red points) throughout the forest site.Measurements in Wytham 2021 were limited due to lack of cloud-free Sentinel-2 images, so we use 2018 data only. 

Each data point is matched with a Sentinel-2 surface reflectance observation. We obtained Level-2A surface reflectance images (10 m resolution) for the dates of each field campaign via the the GEODES Portal which is a digital platform operated by CNES (French National Centre for Space Studies) that serves as a central hub for accessing and utilizing space-based Earth observation data in France\footnote{S2 data portal: \url{https://geodes-portal.cnes.fr/}}. We used images within close day of in-situ measurement when available. For Barrax, we used images on 2018-06-13, 2021-07-22 for 2018 and 2021. For Wytham, clear images on 2018-06-29 were used. We averaged the reflectance of all 10 m pixels falling within each 20×20 m to get one spectrum per ground measurement. The observation angles (sun zenith ~25\textdegree–30\textdegree for Barrax midday acquisitions, view zenith ~0\textdegree since Sentinel-2 is near-nadir) were recorded as inputs. Because FRM4Veg also provided uncertainty estimates for each measurement (e.g., standard deviation of LAI), we consider those in interpreting results, though our algorithm currently does not take measurement uncertainty as input.

\subsubsection{BelSAR}
\begin{figure}[htp]
	\begin{center}
		\centerline{\includegraphics[width=\columnwidth]{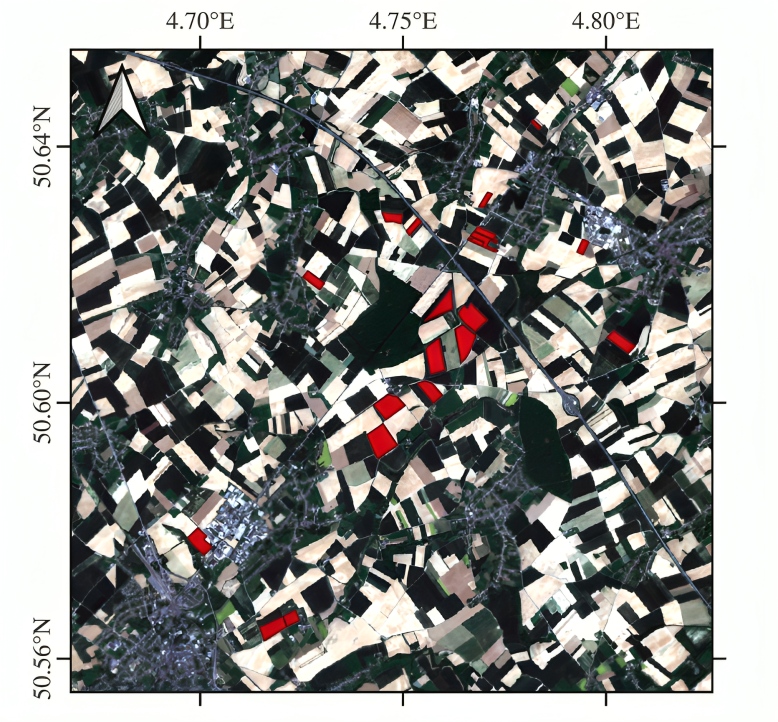}}
		\caption{Agricultural test site used for the BelSAR campaign in central Belgium, showing a mosaic of crop fields with validation sites highlighted in red.}
		\label{fig:belsar}
	\end{center}
\end{figure}

The BelSAR 2018 campaign is an airborne SAR (Synthetic Aperture Radar) experiment in Belgium, which also collected synchronous in-situ vegetation data for validation~\cite{orban2021belsar,bouchat2024belsar}. The test site consists of agricultural fields (primarily winter wheat and maize). Although the project’s focus was on SAR, field teams measured vegetation indices including the Plant Area Index (PAI) for wheat and the Green Area Index (GAI) for maize at multiple dates over the growing season. PAI and GAI are equivalent to LAI~\cite{fang2019overview} for fully green crops (PAI includes all photosynthetic material) – we thus interpret both as LAI in our study. Ten fields of maize and ten fields of winter wheat were instrumented, each $>1$ ha in size. For each field, LAI measurements were taken at 3 or more dates from June to August 2018, covering crop development and senescence. Approximately 3 sample points per field were measured and averaged to give a field-level LAI per date. Figure~\eqref{fig:belsar} shows the agricultural test site used for the BelSAR campaign, showing a mosaic of crop fields with validation sites highlighted in red. No direct chlorophyll measurements were reported for BelSAR, so we evaluate only LAI for this dataset.

We obtained Sentinel-2 reflectance for the BelSAR fields on dates closest to the field campaigns. Based on BelSAR’s timeline, usable Sentinel-2 images were available for 2018-05-18, 2018-05-28, 2018-06-20, 2018-07-15, 2018-07-27 and 2018-08-04 (cloud-free observations within close days of measurement dates). We matched each in-situ data point (field, date) with the Sentinel-2 image closest in time (within 24 days as per the criteria in \cite{zerah2024physics}. This averaging mitigates any geolocation mismatch or within-field variability, yielding one reflectance vector per field per date, along with the corresponding mean LAI. The sun angles for these summer acquisitions were ~30°–35° zenith, with view zenith ~4° (since Belgium is off-nadir in the Sentinel-2 swath) – we input these angles to our network. After filtering out one date with no nearby Sentinel image (Aug 29, 2018), the BelSAR evaluation set contains ~60 samples (20 fields × 3 dates average, each with a measured LAI). 

In total, the complete data contains 211 LAI and 121 CCC reference measurements. FRM4Veg and BelSAR provide a robust test of our model on real data: FRM4Veg covers diverse crop types and a forest with both LAI and CCC ground truth, while BelSAR adds multi-temporal crop data with independent measurement protocols. These datasets also allow comparison to prior work of~\cite{zerah2024physics} evaluated their PROSAIL-VAE and SNAP on these same campaigns, so we can benchmark our results against their reported accuracies.

\section{Results and Comparison}
\label{results}
We evaluated the performance of the Transformer-VAE by comparing predicted LAI and CCC against in-situ values for the FRM4Veg and BelSAR datasets. We report error metrics including the root-mean-square error (RMSE) and the coefficient of determination (R²) between predictions and measurements. We also compare qualitatively to the performance of the Sentinel-2 Biophysical Processor (SNAP BP) and the physics-constrained PROSAIL-VAE method from literature, using reported metrics as references~\cite{weisss2toolbox,zerah2024physics}.

\begin{figure}[h!]
	\vskip 0.2in
	\begin{center}
		\centerline{\includegraphics[width=\columnwidth]{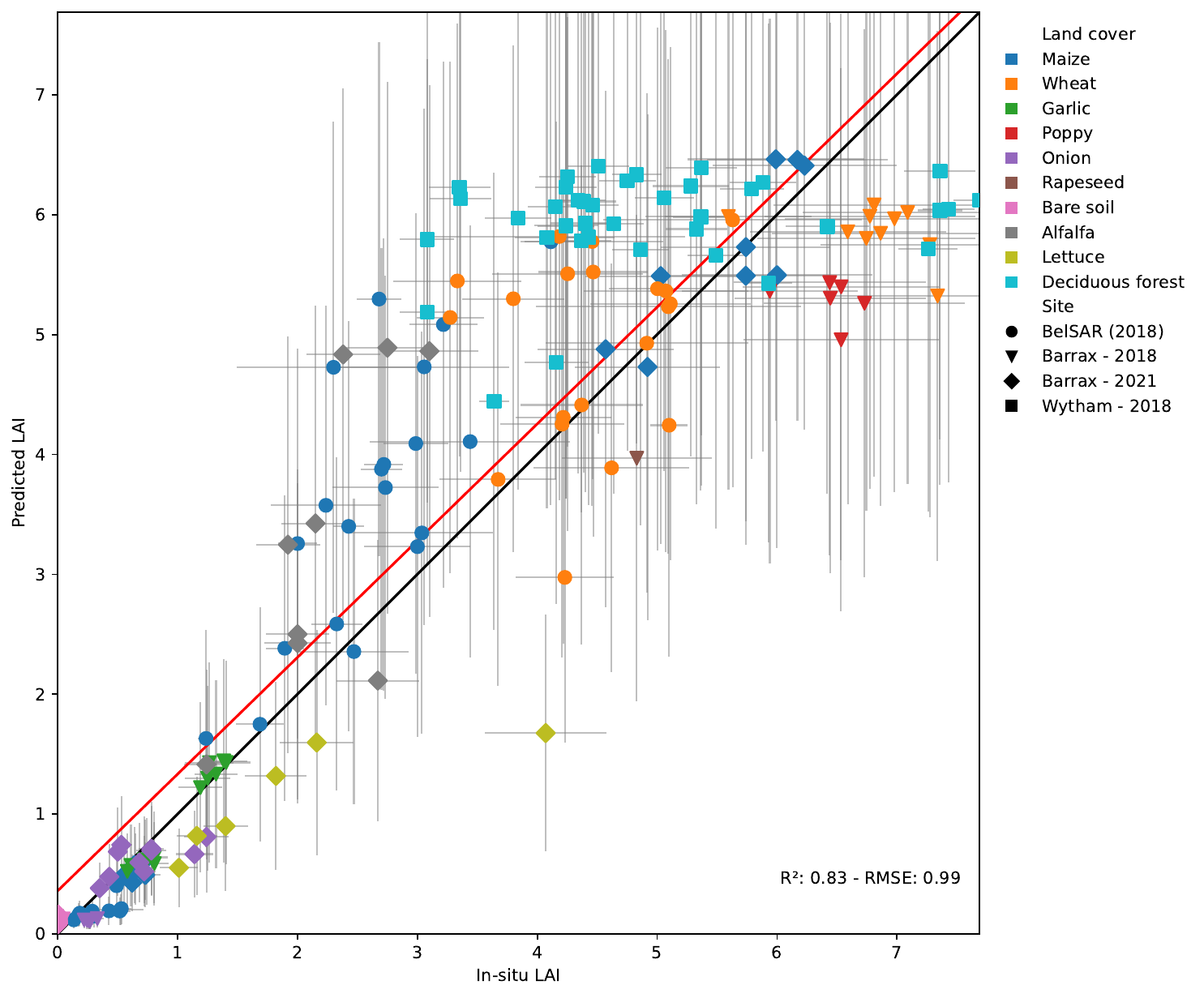}}
		\centerline{\includegraphics[width=\columnwidth]{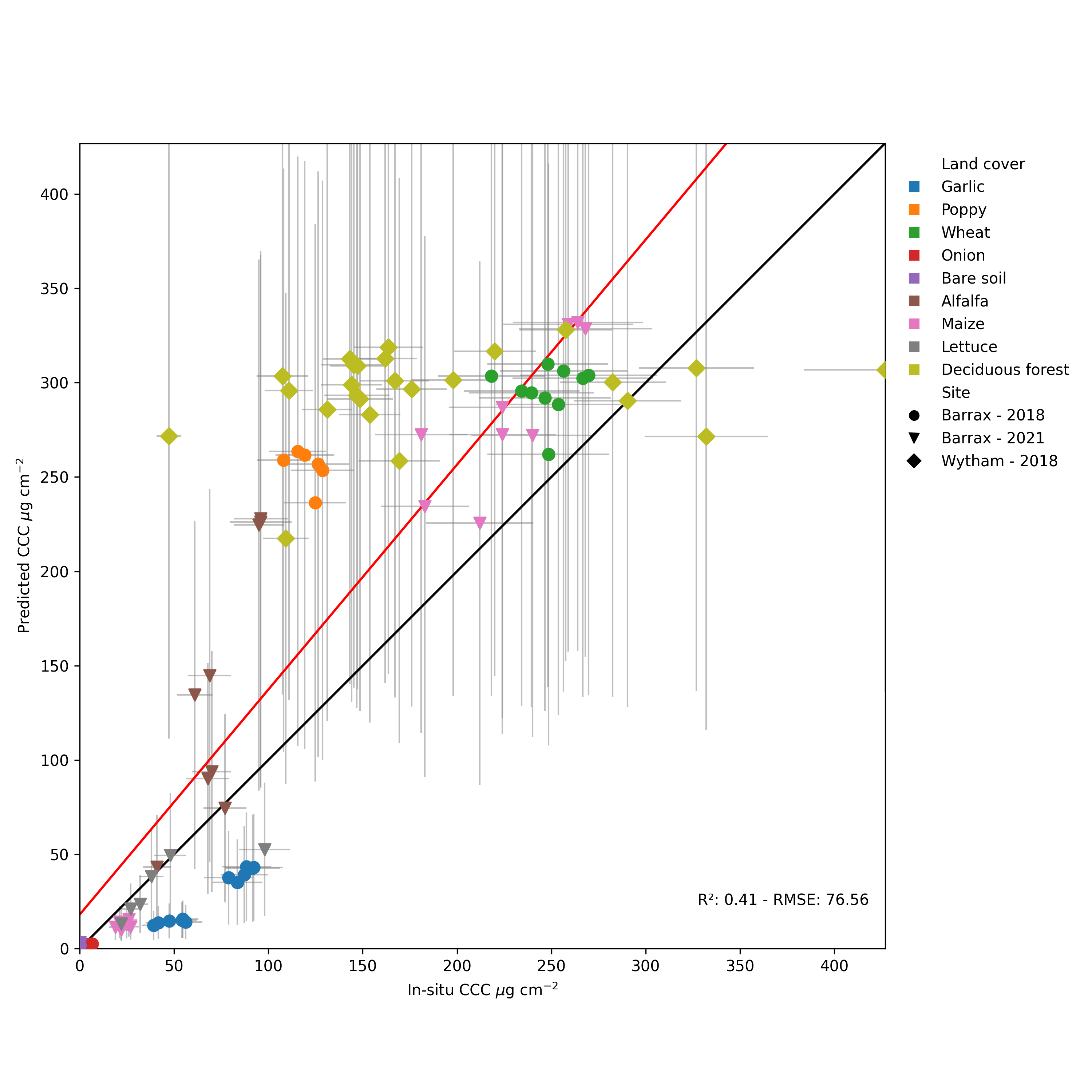}}
		\caption{\textbf{Top:} Predicted versus in-situ measured Leaf Area Index (LAI) across all validation sites and land cover types. \textbf{Bottom:} Predicted versus in-situ measured Canopy Chlorophyll Content (CCC) for the same sites and land covers. Each point represents a field measurement, colored by land cover class and shaped by site. Error bars indicate prediction uncertainty. The black line is the 1:1 line (perfect agreement), and the red line is the best-fit regression.}
		\label{fig:lai_and_ccc}
	\end{center}
	\vskip -0.2in
\end{figure}

On the FRM4Veg dataset (Barrax crops and Wytham forest), and the BelSAR agricultural dataset (winter wheat and maize fields over time), our model’s LAI predictions showed a strong agreement with ground truth across the full range of values. Figure~\eqref{fig:lai_and_ccc} \textbf{Top} illustrates the correlation between predicted and measured LAI for all samples. 

The points cluster close to the 1:1 line, with an overall RMSE of 0.99 LAI units and $R^2$ of 0.83. This accuracy is better than the SNAP biophysical processor, and the PROSAIL-VAE methods which had an RMSE of 1.24 and 1.16, $R^2$ of 0.71 and 0.75 on the same compilation of sites. Our approach, despite using only simulated training, achieved better LAI accuracy to SNAP’s operational algorithm and the PROSAIL-VAE approach.

Notably, our model handles both crop and forest cases in one unified model without site-specific tuning. The largest LAI errors in our results occurred at the high end (LAI $> $5 in the Wytham forest site), where a slight overestimation tendency was observed (several points around In-situ LAI $\sim$4–5 had predicted LAI $\sim$5–6, as seen in Figure~\eqref{fig:lai_and_ccc} \textbf{Top}). Our simulation-trained model did not suffer particular degradation on BelSAR, indicating robust generalization across crop types and phenological stages. Overall the model captured the temporal trends. For instance, fields measured with LAI ~4.5 at peak were predicted ~4.0–4.3; fields with LAI $\sim$1 early in season were predicted ~1.5. These differences might be partly due to the model’s prior which was not tuned specifically to crop phenology. Nonetheless, the predictions are within a reasonable range for agronomic applications (e.g., distinguishing low, medium, high LAI fields).

A major advantage of our multi-output approach is the ability to retrieve canopy chlorophyll content (CCC) accurately by jointly modeling LAI and Cab. Figure~\eqref{fig:lai_and_ccc} \textbf{Bottom} shows the results for CCC on the FRM4Veg data (note: BelSAR lacked CCC ground truth). Our model achieves a low RMSE of 76.56 (in units of $\mu$g $cm^{–2}$ of chlorophyll) and a low $R$ of 0.41. This is a significant improvement over the SNAP biophysical processor, which had reported CCC RMSE on the order of 88 ($\mu$g $cm^{–2}$) and $R^2$ of 0.22 over these same test sites. The error reduction by nearly a factor of two for CCC can be attributed to the physical coupling of LAI and Cab in our model – by retrieving both consistently, we avoid the error amplification that occurs when multiplying separate LAI and Cab estimates as SNAP does (SNAP’s CCC is essentially derived from two independent neural nets for LAI and chlorophyll, which compounds their errors)~\cite{weiss2020s2toolbox}.

Our model achieved less performance compared to the PROSAIL-VAE, which also had CCC RMSE of 42.33 and $R^2$ of 0.82~\cite{zerah2024physics}.  Figure~\eqref{fig:lai_and_ccc} \textbf{Bottom} confirms that our simulation-trained model CCC predictions  are very tightly correlated with In-situ values across the range (~0 to 350). We do observe a few outliers where CCC is slightly overestimated (points around measured 200–250 with predicted 250–300) and underestimated (In-situ $\sim$100 predicted $\sim$50), but these are rare. The model’s ability to retrieve CCC is especially noteworthy given that many training samples were generated with independent Cab and LAI – the network learned to correlate them to match real spectra, effectively learning the concept of CCC. This demonstrates the \textbf{novelty of our simulation-only training:} even without seeing real data, the model inferred the relationship needed to predict derived variables like CCC. It also highlights the benefit of the probabilistic latent space – the model can output a distribution for LAI and Cab such that CCC (CCC=LAI×Cab) is consistent; indeed, we could compute an implied distribution for CCC from the joint samples of LAI and Cab in the decoder. Overall, our model provides physically consistent multi-parameter retrieval, capturing the coupled nature of LAI and CCC (and potentially other variables like leaf dry matter or water content, though those were not validated due to lack of ground truth).

Table~\eqref{tab:comparison} summarizes the performance of our Transformer-VAE (T-VAE) model against literature benchmarks on in-situ validation data across multiple sites. For all sites combined, our model achieved an LAI RMSE of 0.99 and a CCC RMSE of 76.56. In comparison, the SNAP BP method had LAI RMSE of 1.24 and CCC RMSE of 88.08, while the physics-constrained PROSAIL-VAE~\cite{zerah2024physics} achieved LAI RMSE of 1.16 and CCC RMSE of 42.33 on the same dataset.

Thus, our model outperforms both SNAP and PROSAIL-VAE for LAI estimation, but its CCC predictions are less accurate than PROSAIL-VAE, likely due to the compounded uncertainty in estimating this derived variable. Notably, our model was trained purely on simulated data, yet achieves competitive performance with methods trained on real satellite imagery, highlighting the effectiveness of a well-designed simulation and physics-informed approach.

Additionally, our model provides uncertainty estimates for each prediction via the output distribution. For example, the mean prediction interval width (MPIW) for LAI across all sites was 5.15, and the prediction interval coverage probability (PICP) was 0.95, indicating that the true value fell within the model’s predicted 95\% interval about 95\% of the time. These probabilistic outputs are a key advantage over deterministic methods like SNAP, enabling more robust and interpretable biophysical parameter retrieval.
 
\begin{table*}[h!]
	\renewcommand{\arraystretch}{1.3}
	\caption{LAI and CCC prediction performance on in‐situ data for SNAP, PROSAIL‐VAE and Transformer-VAE.\label{tab:perf}}
	\tabcolsep=0pt
	\label{tab:comparison}
	\begin{tabular*}{\textwidth}{@{\extracolsep{\fill}}ll rrrrr rrrr @{\extracolsep{\fill}}}
		\toprule
		& \multicolumn{5}{@{}c@{}}{LAI} & \multicolumn{4}{@{}c@{}}{CCC} \\
		\cline{3-7}\cline{8-11}
		Method         & Metric & BelSAR & Barrax (2018) & Barrax (2021) & Wytham & All
		& Barrax (2018) & Barrax (2021) & Wytham & All \\
		\midrule
		SNAP           & RMSE   & 1.22   & 1.43          & 0.48          & 1.77   & 1.24
		& 83.92         & 84.53         & 101.35 & 88.08 \\
		\toprule
		PROSAIL‐VAE    & RMSE   & 1.30   & 1.42          & 0.72          & 1.21   & 1.16
		& 27.60         & 20.51         & 80.78  & 42.33 \\
		& MPIW   & 4.74   & 3.74          & 2.72          & 5.45   & 4.04
		& 140.53        & 94.02         & 235.20 & 138.18 \\
		& PICP   & 1.00   & 0.88          & 0.96          & 0.95   & 0.95
		& 0.95          & 0.98          & 0.84   & 0.94 \\
		\toprule
		\textbf{T‐VAE}    & RMSE   & \textbf{1.03}   & \textbf{0.72}  & \textbf{0.63}  & 1.53   & \textbf{0.99}
		& 63.43  & 41.56 & 134.75  & 76.56 \\
		& MPIW   & 6.04   & \textbf{1.52}  & \textbf{0.72}  & \textbf{0.96}   & 5.15
		& \textbf{56.09}  & \textbf{30.11} & \textbf{77.89} & 310.55 \\
		& PICP   & 0.92   & 0.67  & 0.37  & 0.10   & 0.95
		& 0.21   & 0.19  & 0.16   & 0.88 \\
		\bottomrule
	\end{tabular*}
\end{table*}
 
 It is important to note that our model was not exposed to any real data during training – all the above results were achieved zero-shot, using knowledge gained from synthetic data and the embedded physics. The fact that we rival SNAP (which was itself calibrated on a large LUT of PROSAIL simulations carefully tuned to real data distributions) and the fact that we match the more complex self-supervised approach, underscores the novelty and effectiveness of the simulation-only training approach. It challenges the notion that simulation-trained models cannot generalize: we have shown that with physics-informed design and appropriate simulation of training data, the gap can be minimized. In our case, the encoder’s flexibility (Transformer layers) may have helped it adjust to subtle differences in real spectra that weren’t perfectly represented in simulation, while the PROSAIL decoder ensured physical consistency.

 \section{Discussion}
\label{discussion}
Our results demonstrate that a physics-informed deep learning model trained exclusively on synthetic data can achieve state-of-the-art accuracy in retrieving biophysical parameters from real satellite images.

A key concern addressed in earlier work was that models trained on radiative transfer simulations might not transfer well to real imagery~\cite{combal2003retrieval,atzberger2004object}. In our study, we tackled these issues by using a transformer-VAE approach with the PROSAIL model and by aligning the simulation parameter space with real-world knowledge (including variable correlations). The high accuracy on field data indicates that our synthetic dataset was representative enough of the real conditions. In essence, we bypassed the need for real imagery in training by ensuring the training simulations were physically comprehensive. 
By integrating PROSAIL as the decoder, our model inherits the strengths of physical inversion (guaranteed physical feasibility of outputs) while leveraging the power of deep learning to handle noise and nonlinear feature extraction. 


During inference, it yields not just point estimates but an uncertainty distribution (via latent variance) for each parameter – a feature that could be extremely useful for assimilation into crop or climate models. In terms of architecture, the choice of a Transformer encoder (as opposed to a simple multi-layer perceptron) appears to have helped capture subtle relationships in the spectral data (especially interactions between bands). We noticed during training that the Transformer-based model converged to a lower reconstruction error than a comparable fully-connected encoder, and it was more stable in predicting extremes (possibly due to better handling of variations in spectral shape such as the red edge position, which might correlate with CCC). The hybrid self-supervised approach of~\cite{zerah2024physics} uses real satellite images to train the model and was shown to outperform pure simulation-trained ones. Our work shows that if one can incorporate enough physics and carefully simulate training data, the gap can be narrowed significantly.

\section{Conclusion}
\label{conclusion}
We presented a Transformer-VAE approach for biophysical parameter estimation that integrates the PROSAIL radiative transfer model into a deep learning framework. The model was trained using only simulated reflectance data, yet it demonstrated excellent performance in retrieving LAI and CCC from Sentinel-2 imagery of real agricultural and forest sites. By leveraging physical knowledge through a differentiable PROSAIL decoder and carefully designing the training simulations (using realistic parameter distributions and correlations), we overcame the traditional limitations of simulation-trained models. 

Our results on FRM4Veg and BelSAR field datasets show that the simulation-only model can achieve accuracies equivalent to state-of-the-art hybrid methods that use real imagery for training, and even outperform the operational SNAP algorithm. The model retrieves multiple canopy and leaf variables simultaneously in a physically consistent way, providing not only point estimates but also uncertainty quantification for each prediction. This work highlights that physics-informed neural networks can serve as powerful tools for remote sensing inversion, effectively blending the strengths of physical and statistical approaches. The novelty lies in demonstrating that with the right constraints and training strategy, real-world generalization is possible without direct real-data training. It’s an encouraging result for Earth observation: as RTMs improve and computing allows training on huge synthetic datasets, we might rely less on expensive field data while still obtaining trustworthy models. This can greatly simplify the development of global bio-physical products, as it reduces dependence on scarce calibration data. In future work, we plan to extend this framework to hyperspectral data with other radiative transfer models to further improve retrieval robustness.

\bibliography{example_paper}
\bibliographystyle{icml2025}

\newpage
\appendix
\onecolumn
%
%

\end{document}